\documentclass{article}
\usepackage[table,xcdraw]{xcolor}

\usepackage{amsmath,amsfonts,bm}

\def\eqref#1{equation~\ref{#1}}

\def\1{\bm{1}}

\DeclareMathAlphabet{\mathsfit}{\encodingdefault}{\sfdefault}{m}{sl}
\SetMathAlphabet{\mathsfit}{bold}{\encodingdefault}{\sfdefault}{bx}{n}

\usepackage[utf8]{inputenc} 
\usepackage[T1]{fontenc}    
\usepackage{hyperref}       
\usepackage{url}            
\usepackage{booktabs}       
\usepackage{amsfonts}       
\usepackage{amsmath, bm, amsthm}
\usepackage{mathtools}
\usepackage{nicefrac}       
\usepackage{microtype}      
\usepackage{graphicx,wrapfig}
\usepackage{subcaption}
\usepackage{tabularray}
\usepackage{listings}
\usepackage{hhline}
\usepackage{listings}
\usepackage{multirow, amssymb}
\usepackage[normalem]{ulem}
\useunder{\uline}{\ul}{}

\usepackage{algorithm} 
\usepackage{algorithmic}
\newcounter{numquote}
\usepackage{enumitem}
\usepackage{mathtools}

\definecolor{headers}{RGB}{64,64,64}
\definecolor{cellcol}{RGB}{217,217,217}
\definecolor{first}{RGB}{198,247,198}
\definecolor{second}{RGB}{255,219,176}

\newcommand{\name}{PAGER}

\usepackage{hyperref}
\usepackage[accepted]{icml2024}

\usepackage[capitalize,noabbrev]{cleveref}

\theoremstyle{plain}

\theoremstyle{definition}

\theoremstyle{remark}

\usepackage[textsize=tiny]{todonotes}

\icmltitlerunning{PAGER: Accurate Failure Characterization in Deep Regression Models}

\begin{document}

\twocolumn[
\icmltitle{PAGER: Accurate Failure Characterization in Deep Regression Models}

\icmlsetsymbol{equal}{*}

\begin{icmlauthorlist}
\icmlauthor{Jayaraman J. Thiagarajan}{yyy}
\icmlauthor{Vivek Narayanaswamy}{yyy}
\icmlauthor{Puja Trivedi}{zzz}
\icmlauthor{Rushil Anirudh}{aaa}

\end{icmlauthorlist}

\icmlaffiliation{yyy}{Lawrence Livermore National Labs, CA, USA}
\icmlaffiliation{zzz}{University of Michigan, USA}
\icmlaffiliation{aaa}{Amazon, CA, USA}

\icmlcorrespondingauthor{Jayaraman J. Thiagarajan}{jjayaram@llnl.gov}

\icmlkeywords{failure detection, deep neural networks, uncertainty quantification, risk prediction, extrapolation}

\vskip 0.3in

\vskip 0.3in
]

\printAffiliationsAndNotice{}

\begin{abstract}
Safe deployment of AI models requires proactive detection of failures to prevent costly errors. To this end, we study the important problem of detecting failures in deep regression models. Existing approaches rely on epistemic uncertainty estimates or inconsistency w.r.t the training data to identify failure. Interestingly, we find that while uncertainties are necessary they are insufficient to accurately characterize failure in practice. Hence, we introduce PAGER (Principled Analysis of Generalization Errors in Regressors), a framework to systematically detect and characterize failures in deep regressors. Built upon the principle of anchored training in deep models, PAGER unifies both epistemic uncertainty and complementary manifold non-conformity scores to accurately organize samples into different risk regimes.

\end{abstract}

\section{Introduction}
\label{sec:intro}

\begin{figure*}
    \centering
    \includegraphics[width = 0.9\textwidth]{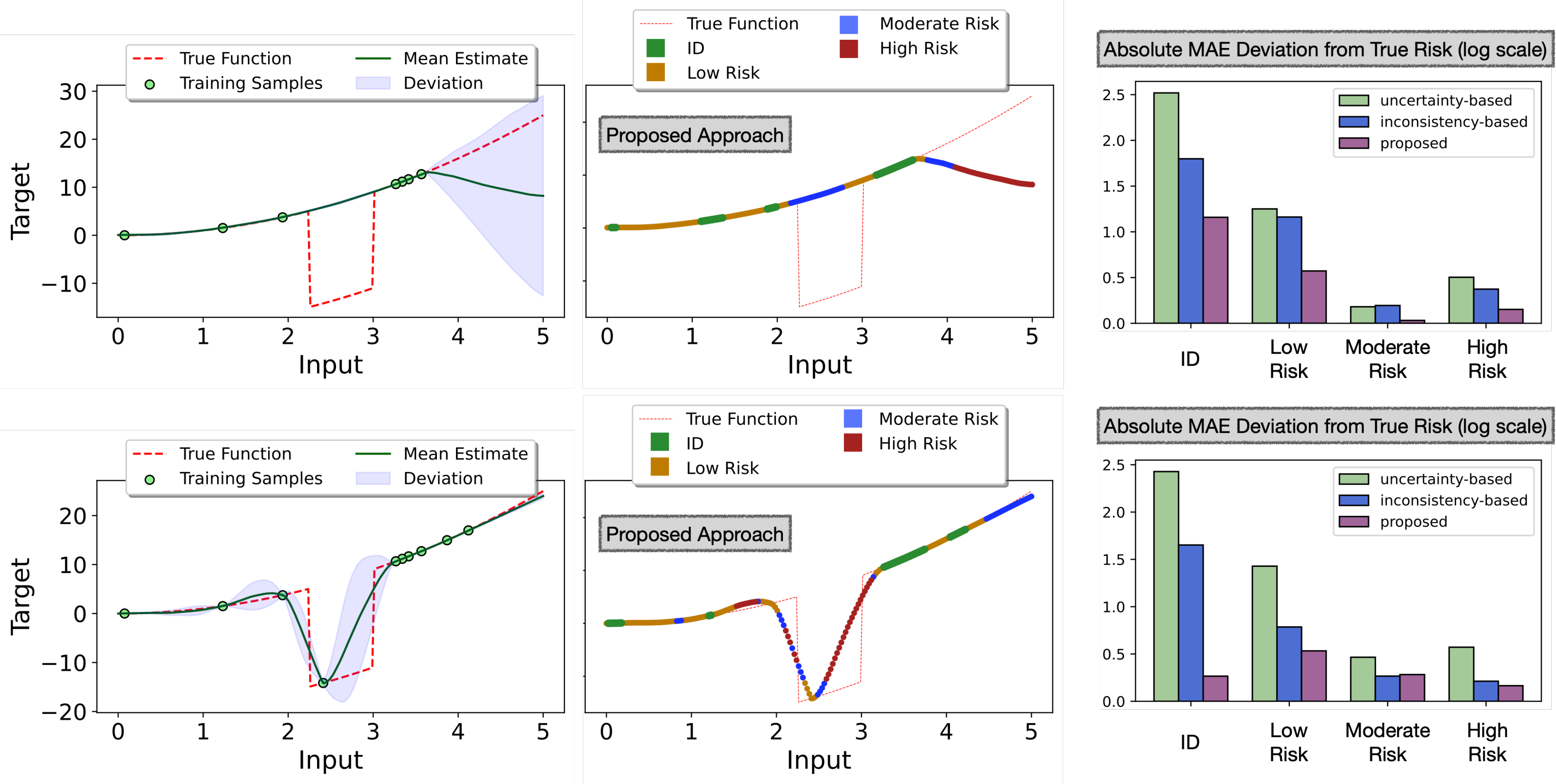}
    \caption{\textbf{Epistemic uncertainty, while necessary, is not sufficient to completely characterize all risk regimes.} \underline{Top:} Out-of-support (OOS) samples in the range of $[2.2-2.7]$ exhibit low uncertainty but moderate risk due to significant deviation from true function. \underline{Bottom:} Even with better experiment designs, uncertainty alone in the extrapolating regime $[4.5-5]$ is unreliable due to potential drift from the truth. We propose \name~, a framework that leverages anchoring \citep{thiagarajan2022single} to unify prediction uncertainty and non-conformity to the training data manifold. \name~~accurately flags those erroneous regimes as Moderate Risk (shown in \textcolor[HTML]{1E2FF5}{blue}) and outperforms existing baselines in accurately categorizing samples consistent with the true risk (lower MAE).}
    \label{fig:teaser}
\end{figure*}

Ensuring the safety of AI models involves proactively detecting failures to help avoid costly errors. While existing efforts have predominantly focused on classification models~\citep{guillory2021predicting,narayanaswamy2022predicting,baek2022agreement}, we are interested more challenging problem of failure detection in deep regressors. Though continuous-valued prediction is prevalent in high-impact applications including healthcare~\citep{luo2022biogpt}, physical sciences~\citep{raissi2019physics} and reinforcement learning, the notion of failure in regression models is often application-specific. 

A common approach is to use epistemic uncertainty~\citep{lakshminarayanan2017simple, gal2016dropout, he2020bayesian, amini2020deep} as a surrogate for expected risk~\citep{deup}. However, we uncover a crucial insight that uncertainties alone are insufficient for a comprehensive characterization of failures in regression models. Figure \ref{fig:teaser} empirically illustrates the lack of correlation between uncertainty and the true risk using a simple $1$D function (with two experiment designs) -- low uncertainty regimes can still correspond to a higher risk due to feature heterogeneity in the training data~\citep{datasuite}, and similarly data regimes outside the training support may correspond to low risk if the model extrapolates well. 

To circumvent this, we introduce \name~({\ul P}rincipled {\ul A}nalysis of {\ul G}eneralization {\ul E}rrors in {\ul R}egressors), a new framework for failure analysis. A key contribution of this work is that we advocate for incorporating manifold non-conformity, \textit{i.e.}, adherence to the joint data distribution, as an essential complement to uncertainties. Building upon the principle of anchored training~\citep{thiagarajan2022single}, we make a critical finding that non-conformity scores can be estimated through \textit{reverse anchoring} without the need for auxiliary models. Additionally, we propose a flexible analysis of model errors through the concept of risk regimes, thus avoiding the need for a rigid definition of failure or additional calibration data. Finally, we introduce a suite of metrics to holistically assess failure detectors in regression tasks. Empirical results reveal that, when compared to state-of-the-art detectors, the risk regimes identified by \name~align best with the true risk. 
\section{Background and Related Work}
\label{sec:back}
\begin{figure*}[t]
    \centering
    \includegraphics[width = 0.83\textwidth]{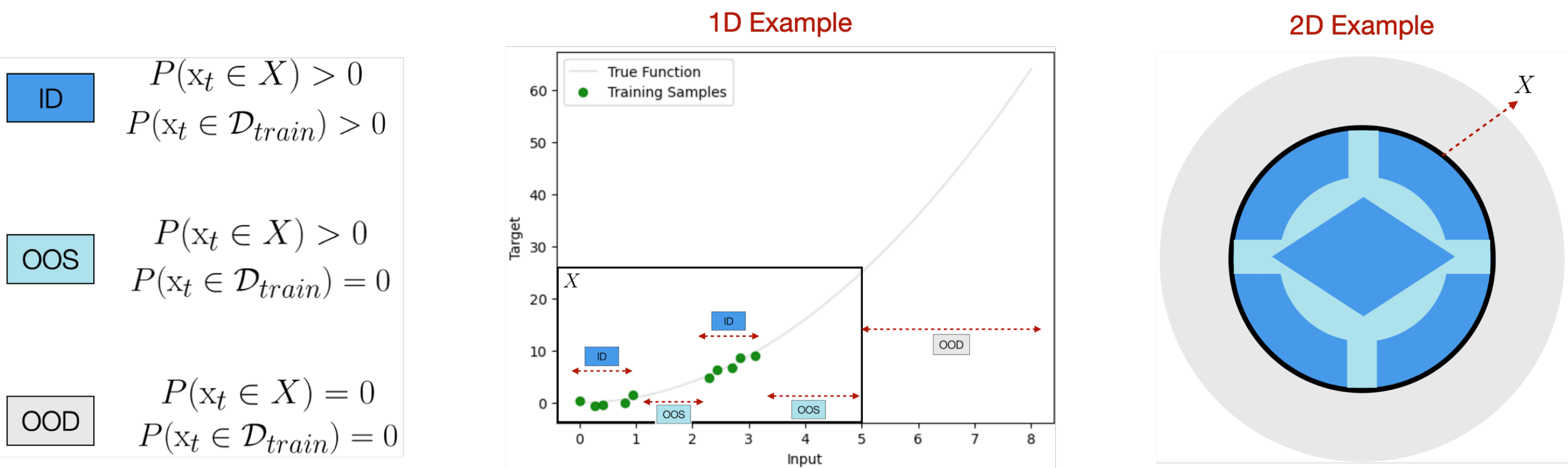}
    \caption{\textbf{An illustration of different risk regimes.} Using examples in 1D and 2D, we show ID, OOS and OOD regimes.}
    \label{fig:oos_ood}
\end{figure*}

\textbf{Preliminaries.} We consider a predictive model $F$, parameterized by $\theta$, trained on a labeled dataset $ \mathcal{D} = \{(\mathrm{x}_i, \mathrm{y}_i)\}^{M}_{i=1}$ with $M$ samples. Each input $\mathrm{x}_{i} \in X$ and label $\mathrm{y}_i \in y$ belong to the spaces of inputs $X$ (in $d-$dimensions) and continuous-valued targets $y$ respectively. Given a non-negative loss function $\mathcal{L}$, e.g., absolute error $|\mathrm{y}-\mathrm{\hat{y}}|$, the sample-level risk of a predictor can be defined as $\mathrm{R}(\mathrm{x}; F_{\theta}) = \mathbb{E}_{y|\mathrm{x}}~{\mathcal{L}(\mathrm{y}, F_{\theta}(\mathrm{x})})$. Since estimating true risk is non-trivial due to the need for access to the unknown joint distribution $P(X, y)$, an alternative is to identify risk regimes in accordance to the expected risk. We now define the different risk regimes that one needs to characterize: (i) \underline{In-distribution (ID)}: This is the scenario where $P(\mathrm{x}_t \in X) > 0$ and $P(\mathrm{x}_t \in \mathcal{D}) > 0$, \textit{i.e.}, there is likelihood for observing the test sample in the training dataset; (ii) \underline{Out-of-Support (OOS)}: The scenario where $P(\mathrm{x}_t \in X) > 0$ but $P(\mathrm{x}_t \in \mathcal{D}) = 0$, \textit{i.e.}, the train and test sets have different supports, even though they are drawn from the same space; (iii) \underline{Out-of-Distribution (OOD)}: This is the scenario where $P(\mathrm{x}_t \in X) = 0$, \textit{i.e.}, the input spaces for train and test data are disjoint. Figure \ref{fig:oos_ood} illustrates the differences between OOS and OOD using 1D and 2D examples. In case of 1D, OOS corresponds to regimes where the likelihood of observing data in the training support is zero but is non-zero in the input-space. Similarly, in 2D, OOS constitutes regimes with new combinations of features (light blue) which are not jointly seen in the train data.

\textbf{Failure Characterization}. In classification tasks, incorrectly assigned labels are considered failures. Hence, failure detectors can estimate either the sample-level \textit{correctness}~\citep{ng2022predicting,jiang2022assessing} or distribution-level scores such as generalization gap~\citep{guillory2021predicting,narayanaswamy2022predicting, chen2021mandoline, jiang2018predicting, deng2021labels}. Risk estimation in regression models is more challenging since the the notion of failure is highly subjective, \textit{i.e.}, permissible tolerance levels on prediction errors can vary across use-cases. Among existing methods, DEUP~\citep{deup} is a recent approach that utilized predictive uncertainty as a surrogate for total risk, which we illustrated to be insufficient for failure detection in Figure ~\ref{fig:teaser}. Conformal prediction (CP) forms another popular class of uncertainty estimation methods~\citep{vovk2005algorithmic, lei2018distribution}, that can be leveraged to identify risk regimes. However, with OOS and OOD data, the exchangeability assumption made by CP frameworks is violated~\citep{tibshirani2019conformal} and hence the estimated intervals can be erroneous. Finally, approaches such as DataSUITE~\citep{datasuite} qualify failure solely based on feature inconsistency with respect to the training data distribution. However, our results show that such methods are incapable of identifying errors in OOS regimes. 

\textbf{Anchoring in Deep Models.} Anchored training involves reparameterizing an input sample $\mathrm{x}$ (referred to as the \textit{query}) into a tuple comprising an \textit{anchor} $\mathrm{r}$ randomly drawn from the training data and the residual $\Delta \mathrm{x}$ denoted by $[\mathrm{r}, \Delta \mathrm{x}] = [\mathrm{r}, \mathrm{x}-\mathrm{r}]$~\citep{thiagarajan2022single}. It induces a joint distribution that depends not only on $P(X)$, but also on the distribution of residuals $P(\Delta)$. In practice, the sole modification lies in the input layer, requiring additional dimensions for vector-valued data or channels for images, and a modest addition of parameters to the first layer. This approach is adaptable to any architecture (e.g., MLP, CNN, ViT). During training, we enforce consistency in predictions for a query $\mathrm{x}$ across all possible anchors. Consequently, at inference time, one can obtain predictions and corresponding uncertainties by marginalizing out the anchor choice. A detailed description can be found in Appendix \ref{app: 1}.

\section{Failure Detection in Deep Regressors}
\label{sec:approach}

The central idea of our approach is to obtain not only uncertainty estimates, but also manifold non-conformity (MNC) scores, for failure detection. This is motivated by the observation that, regardless of the uncertainty, a model can induce a large error for test sample $\mathrm{x}_t$, when $(\mathrm{x}_t, \mathrm{y}_t) \notin P(X,y)$, \textit{i.e.}, the risk can be high when the sample does not adhere to the data manifold. While there exist several approaches for estimating epistemic uncertainty~\citep{gawlikowski2023survey, yang2021generalized}, measuring non-conformity without ground truth is not straightforward. Consequently, it is typical to adopt scoring functions only based on inputs~\citep{datasuite} or utilize CP strategies to transform scores into calibrated intervals~\citep{fcp}. While the former approach does not leverage the characteristics of the task, the latter is not applicable in our scenario due to the violation of the exchangeability condition w.r.t OOS and OOD regimes. Hence, we propose an alternative approach based on anchored neural networks.

\begin{figure*}[t]
    \centering
    \includegraphics[width = 0.8\textwidth]{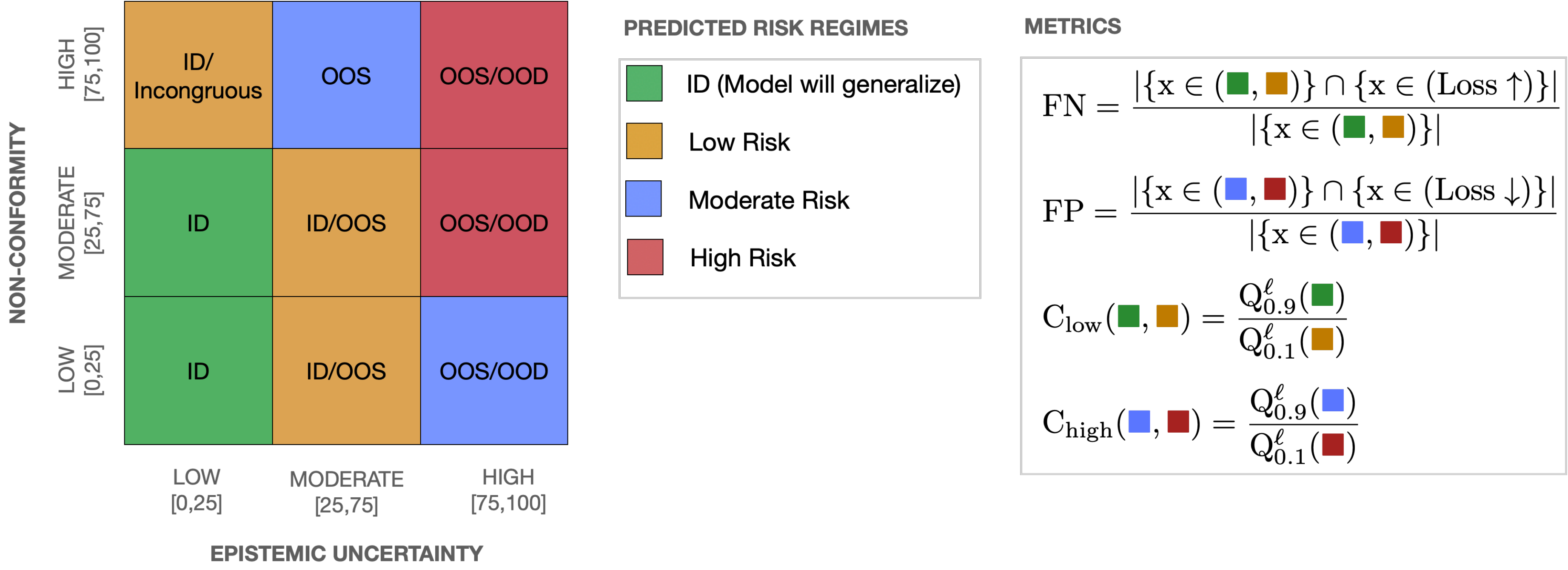}
    \caption{\textbf{Overview of our proposed framework.}~\name~organizes test examples into bins (\textit{low}, \textit{moderate} and \textit{high}) using both predictive uncertainty and MNC scores. With such a categorization, \name~associates samples into $4$ levels of expected risk (ID, Low Risk, Moderate Risk and High Risk). We also advocate a suite of metrics that enables a holistic assessment of failure detectors.}
    \label{fig:framework}
\end{figure*}

\textbf{Uncertainty via Forward Anchoring}. An anchored model is trained by transforming a sample $\mathrm{x}$ into a tuple, $[\mathrm{r}, \mathrm{x} - \mathrm{r}]$ based on an anchor $\mathrm{r}$, which is also drawn randomly from training data. Building upon findings from~\citep{thiagarajan2022single}, multiple anchors can be used to obtain the mean predictions and uncertainties for a test sample as follows:
\begin{align}
    \label{eqn:deluq_pred}
    \nonumber &\mu(y_t|\mathrm{x}_t) = \frac{1}{K}\sum_{k=1}^K F([\mathrm{r}_k,\mathrm{x}_t-\mathrm{r}_k] ); \\ &\sigma(y_t|\mathrm{x}_t) = \sqrt{\frac{1}{K-1}\sum_{k=1}^K (F([\mathrm{r}_k,\mathrm{x}_t-\mathrm{r}_k]) - \mu)^2 },
\end{align}where $\mu$ and $\sigma$ are estimated by marginalizing across $K$ anchors $\{\mathrm{r}_k\}_{k=1}^{K}$ sampled from $\mathcal{D}$.

\textbf{Non-conformity via Reverse Anchoring}. Turning our attention to the assessment of non-conformity, we make a noteworthy observation regarding the flexibility of an anchored neural network. It is not only able to capture the relative representation of a query in relation to an anchor, but also the reverse scenario. To elaborate, for a given test sample $\mathrm{x}_t$, we swap the roles of query $\mathrm{x}_t$ and anchor $r$ to obtain the prediction for the anchor as $F([\mathrm{x}_t, \mathrm{r} - \mathrm{x}_t])$. Since the ground truth function value is known for the training samples, we can measure the non-conformity score for a query based on its ability to accurately recover the target of the anchor. Note, unlike existing approaches, this can be directly applied to unlabeled test samples and does not require explicit calibration.

Looking from another perspective, the original `anchor-centric' model provides reliable predictions for an input $[\mathrm{r}, \Delta]$ only when $\mathrm{r} \in \mathcal{D}$ and $\Delta \in P(\Delta)$. However, for OOS or OOD samples, if $\Delta \notin P(\Delta)$, the estimated uncertainty becomes large everywhere, and thus becomes inherently unreliable to rank them by levels of expected risk. In contrast, the proposed `query-centric' score overcomes this by directly measuring the discrepancy with respect to the ground truth target. We define our MNC score as follows:
\begin{equation}
\mathtt{Score_1}(\mathrm{x}) = \max_{\mathrm{r} \in \mathcal{D}} \bigg\|\mathrm{y}_{\mathrm{r}} - F([\mathrm{x}, \mathrm{r} - \mathrm{x}])\bigg\|_1
\label{eq:score1}
\end{equation}Note that, we measure the largest discrepancy across the training dataset. In practice, this can be done for a small batch of randomly chosen training samples (e.g., 100). 

\textbf{Resolving Moderate and High Risk Regimes Better.} A closer look of \eqref{eq:score1} reveals that for samples that are far away from the training manifold, the model prediction can be uniformly bad (i.e., extrapolation), as both $\mathrm{x} \notin \mathcal{D}$ and $\Delta \notin P(\Delta)$. This can make distinguishing between samples with moderate and high risk challenging. To mitigate this, we propose to transform both the query $\mathrm{x}$ (used as the anchor in reverse anchoring) and $\Delta$ to be in-distribution so that the anchored model $F$ can produce reliable predictions. We achieve this using the following optimization problem:
\begin{align}
\begin{split}
\mathtt{Score_2}(\mathrm{x}) = \max_{r \in \mathcal{D}}  \bigg\|\mathrm{x} - \arg \min_{\bar{\mathrm{x}}} & \bigg(\bigg\|\mathrm{y}_{\mathrm{r}} - F([\bar{\mathrm{x}}, \mathrm{r} - \bar{\mathrm{x}}])\bigg\|_1 + \\ &\lambda \mathcal{R}(\bar{\mathrm{x}}) \bigg)\bigg\|_2, 
\end{split}
\label{eq:score2}
\end{align}
where $\mathcal{R}(\bar{\mathrm{x}}) = \bigg\|\bar{\mathrm{x}} - A([\mathrm{x}, \bar{\mathrm{x}} - \mathrm{x}])\bigg\|_2 + \bigg\|\mathrm{x} - A([\bar{\mathrm{x}}, \mathrm{x} - \bar{\mathrm{x}}])\bigg\|_2$. In this approach, the score is measured as the discrepancy in the input space to a new fictitious sample that serves as an intermediate anchor, such that its prediction matches the known prediction on the training sample. In other words, we optimize the modification of the query sample $\mathrm{x}$ to $\bar{\mathrm{x}}$ in such a way that we accurately match the true target for the anchor $\mathrm{r}$.
The MNC is then quantified as the amount of movement required in $\mathrm{x}$ to match the target. To ensure that the resulting $\bar{\mathrm{x}}$ remains within the input data manifold, we incorporate a regularizer $\mathcal{R}(\bar{\mathrm{x}})$. Specifically, we train an anchored auto-encoder $A$ on the training dataset $\mathcal{D}$ and enforce cyclical consistency, where $A$ is required to recover $\mathrm{x}$ using $\bar{\mathrm{x}}$ as the anchor and vice versa. We provide the algorithm listings and details of all these methods in Appendix \ref{app: 2}.  


\begin{table*}[t]
\centering
\caption{\textbf{Metrics for 1D Benchmarks.} We report the FN, FP, C$_{\text{low}}$ and C$_{\text{high}}$ metrics on evaluation data across the entire target regime (lower the better). Note that for every metric, we identify the \textcolor{red}{first} and \textcolor{blue}{second} best approach across the different benchmarks.}
\renewcommand{\arraystretch}{1.15}
\resizebox{\textwidth}{!}{
\begin{tabular}{|
>{\columncolor[HTML]{C0C0C0}}c |
>{\columncolor[HTML]{EFEFEF}}c |c|c|c|c|
|
>{\columncolor[HTML]{C0C0C0}}c |
>{\columncolor[HTML]{EFEFEF}}c |c|c|c|c|}
\hline
\textbf{Metric}                                                            & \cellcolor[HTML]{C0C0C0}\textbf{Method} & \cellcolor[HTML]{C0C0C0}\textbf{$f_1(\mathrm{x})$} & \cellcolor[HTML]{C0C0C0}\textbf{$f_2(\mathrm{x})$}  & \cellcolor[HTML]{C0C0C0}\textbf{$f_3(\mathrm{x})$}  & \cellcolor[HTML]{C0C0C0}\textbf{$f_4(\mathrm{x})$}&\textbf{Metric}                                                            & \cellcolor[HTML]{C0C0C0}\textbf{Method} & \cellcolor[HTML]{C0C0C0}\textbf{$f_1(\mathrm{x})$} & \cellcolor[HTML]{C0C0C0}\textbf{$f_2(\mathrm{x})$}  & \cellcolor[HTML]{C0C0C0}\textbf{$f_3(\mathrm{x})$}  & \cellcolor[HTML]{C0C0C0}\textbf{$f_4(\mathrm{x})$} \\ \hline \hline
\cellcolor[HTML]{C0C0C0}                                                    & DEUP                                    & 6.19 & 6.56 & 16.57 & 27.13                                                           &\cellcolor[HTML]{C0C0C0}                                                    & DEUP                                    & 65.90 & 57.86 & 34.13 & 169.54                                                           \\ 
\cellcolor[HTML]{C0C0C0}                                                    & DataSUITE                              & 14.03 & 8.8 & 16.31 & 7.2 &\cellcolor[HTML]{C0C0C0}                                                    & DataSUITE                              & 59.42 & 24.61 & 22.44 & 89.51 \\
 & MNC-only   & 6.19  & 2.26   & 13.73  & 8.84 & & MNC-only   & 57.54 & 40.1  & 31.66 & 52.24    \\
 & Anchor UQ-only & 5.95  & 5.37  & 14.49  & 11.80 & & Anchor UQ-only & 40.7  & 19.88 & 25.59 & 98.92  \\
 \cellcolor[HTML]{C0C0C0}                                                    & \name ~($\mathtt{Score_1}$)                          & \textcolor{blue}{5.61} & \textcolor{red}{0.0} & \textcolor{blue}{11.63} & \textcolor{red}{2.40}                 & \cellcolor[HTML]{C0C0C0}                                                    & \name ~($\mathtt{Score_1}$)                          & \textcolor{blue}{28.08} & \textcolor{red}{7.19} & \textcolor{blue}{19.94} & \textcolor{red}{12.05}                  \\ 
\multirow{-6}{*}{\cellcolor[HTML]{C0C0C0}{FN}$\downarrow$}                       & \name ~($\mathtt{Score_2}$)                          & \textcolor{red}{4.79} & \textcolor{blue}{5.59} & \textcolor{red}{8.43} & \textcolor{blue}{5.59} &\multirow{-6}{*}{\cellcolor[HTML]{C0C0C0}{$\text{C}_{\text{low}}$}$\downarrow$}                       & \name ~($\mathtt{Score_2}$)                          & \textcolor{red}{20.61} & \textcolor{blue}{17.82} & \textcolor{red}{16.57} & \textcolor{blue}{19.74}  \\ \hline \hline    
\cellcolor[HTML]{C0C0C0}                                                    & DEUP                                    & 8.91 & 3.41 & 8.54 & 9.09 &\cellcolor[HTML]{C0C0C0}                                                    & DEUP                                    & 91.64 & \textcolor{blue}{4.47} & 59.46 & 16.56                                                                                                                    \\ 
\cellcolor[HTML]{C0C0C0}                                                    & DataSUITE                              & 18.67 & 15.97 & 19.96 & \textcolor{blue}{5.33} & \cellcolor[HTML]{C0C0C0}                                                    & DataSUITE                              & 3.66 & 46.02 & 58.32 & \textcolor{blue}{6.81} \\

& MNC-only   & 9.93  & 10.42 & 8.82 & 12.18 & & MNC-only   & 33.09 & 18.85 & 29.98 & 20.31 \\
 & Anchor UQ-only & 5.05  & 4.93   & 6.54  & 6.01 & & Anchor UQ-only & 36.05 & 7.75  & 17.92  & 11.56  \\
 
\cellcolor[HTML]{C0C0C0}                                                    & \name ~($\mathtt{Score_1}$)                          & \textcolor{blue}{2.67} & \textcolor{red}{0.0} & \textcolor{blue}{4.67} & 6.67               & \cellcolor[HTML]{C0C0C0}                                                    & \name ~($\mathtt{Score_1}$)                          & \textcolor{red}{3.09} & \textcolor{red}{3.43} & \textcolor{red}{8.78} & 6.88     \\ 
\multirow{-6}{*}{\cellcolor[HTML]{C0C0C0}{FP}$\downarrow$}                       & \name ~($\mathtt{Score_2}$)                          & \textcolor{red}{1.33} & \textcolor{blue}{2.67} & \textcolor{red}{4.33} & \textcolor{red}{4.00} &\multirow{-6}{*}{\cellcolor[HTML]{C0C0C0}{$\text{C}_{\text{high}}$}$\downarrow$}                      & \name ~($\mathtt{Score_2}$)                          & \textcolor{red}{3.09} & 4.67 & \textcolor{blue}{10.99} & \textcolor{red}{5.71}  \\ \hline

\end{tabular}
}
\label{tab:1d}
\end{table*}

\begin{table*}[t]
\centering
\caption{\textbf{Assessing the identified risk regimes for regression benchmarks (Gaps) with dimensionality ranging between $2$ and $32,000$.} We report the FN, FP, C$_{\text{low}}$ and C$_{\text{high}}$ metrics on evaluation data across the entire target regime (lower the better). Note that for every metric, we identify the \textcolor{red}{first} and \textcolor{blue}{second} best approach across the different benchmarks.}
\renewcommand{\arraystretch}{1.15}
\resizebox{0.9\textwidth}{!}{
\begin{tabular}{|
>{\columncolor[HTML]{C0C0C0}}c |
>{\columncolor[HTML]{EFEFEF}}c |c|c|c|c|c|c|c|c|c|}
\hline
                                                          &\cellcolor[HTML]{C0C0C0}&\cellcolor[HTML]{C0C0C0}&\cellcolor[HTML]{C0C0C0}&\cellcolor[HTML]{C0C0C0}&\cellcolor[HTML]{C0C0C0}&\cellcolor[HTML]{C0C0C0}&\cellcolor[HTML]{C0C0C0}&\cellcolor[HTML]{C0C0C0}&\cellcolor[HTML]{C0C0C0}&\cellcolor[HTML]{C0C0C0} \\
\multirow{-2}{*}{\textbf{Metrics}}  & \multirow{-2}{*}{\cellcolor[HTML]{C0C0C0}\textbf{Method}} & \multirow{-2}{*}{ \cellcolor[HTML]{C0C0C0} \textbf{Camel}} & \multirow{-2}{*}{ \cellcolor[HTML]{C0C0C0}\textbf{Levy}} & \multirow{-2}{*}{ \cellcolor[HTML]{C0C0C0}\textbf{Airfoil}} & \multirow{-2}{*}{ \cellcolor[HTML]{C0C0C0}\textbf{NO2}} & \multirow{-2}{*}{ \cellcolor[HTML]{C0C0C0}\textbf{Kinematics}} & \multirow{-2}{*}{ \cellcolor[HTML]{C0C0C0}\textbf{Puma}} & \multirow{-2}{*}{ \cellcolor[HTML]{C0C0C0}\textbf{Housing}} & \multirow{-2}{*}{ \cellcolor[HTML]{C0C0C0}\textbf{Ailerons}} & \multirow{-2}{*}{ \cellcolor[HTML]{C0C0C0}\textbf{DTI}}
                                                          
                                                          \\ \hline \hline
\cellcolor[HTML]{C0C0C0}                                                    & DEUP                                    & 15.79                                       & \textcolor{red}{9.25} &8.81&2.27                             & {\textcolor{blue} {17.58}}                                       & 13.21                                       & 11.46                                          & 14.39             & 16.51                               \\ 
\cellcolor[HTML]{C0C0C0}                                                    & DataSUITE                              & 21.74                                       & 19.69    &5.95&6.58                                  & 18.40                                            & 16.77                                      & 17.71                                          & 11.23                                       &  29.18     \\ 
\cellcolor[HTML]{C0C0C0}                                                    & \name ~($\mathtt{Score_1}$)                          & {\textcolor{blue} {12.15}}                                 & 10.86     & \textcolor{red}{0.75}    & \textcolor{red}{0.0}                                                                & \textcolor{red}{6.42}                                     & \textcolor{red}{10.37}                              & \textcolor{red}{6.25}                                  & \textcolor{red}{0.91} &  \textcolor{red}{9.26}                                                                  \\ 
\multirow{-4}{*}{\cellcolor[HTML]{C0C0C0}{FN}$\downarrow$}                       & \name ~($\mathtt{Score_2}$)                          & \textcolor{red}{11.39}                              & {\textcolor{blue} {10.65}}  & \textcolor{blue}{1.04}                               & \textcolor{blue}{0.93}                                                            & \textcolor{red}{6.38}                                     & {\textcolor{blue} {10.84}}                                 & {\textcolor{blue} {7.29}}                                     & {\textcolor{blue} {1.20}}  &  \textcolor{blue}{10.11}                                                                                      \\ \hline \hline
\cellcolor[HTML]{C0C0C0}                                                    & DEUP                                    & 17.48                                       & 10.04      &6.24&11.79                                & 18.67                                            & 12.05                                       & 10.34                                          & 15.96                                     &  19.73        \\ 
\cellcolor[HTML]{C0C0C0}                                                    & DataSUITE                              & 15.74                                       & 15.32     &6.35&18.33                                 & \textcolor{red}{10.67}                                   & 17.33                                      & 12.07                                          & 8.03                               & 30.93                \\ 
\cellcolor[HTML]{C0C0C0}                                                    & \name ~($\mathtt{Score_1}$)                          & \textcolor{red}{3.36}                               & {\textcolor{blue} {5.04}}  & \textcolor{red}{3.56}                               & \textcolor{blue}{4.18}                                                             & 12.04                                           & {\textcolor{blue} {9.67}}                                 & \textcolor{red}{8.62}                                  & {\textcolor{blue} {4.05}}                               &  \textcolor{red}{9.94}                                      \\ 
\multirow{-4}{*}{\cellcolor[HTML]{C0C0C0}{FP}$\downarrow$}                       & \name ~($\mathtt{Score_2}$)                          & {\textcolor{blue} {7.56}}                                  & \textcolor{red}{4.18}     & \textcolor{blue}{3.82}                               & \textcolor{red}{3.05}                                                        & \textcolor{red}{10.67}                                   & \textcolor{red}{8.83}                              & {\textcolor{blue} {9.07}}                                     & \textcolor{red}{1.33}                            &  \textcolor{blue}{10.29}                                    \\ \hline \hline
\cellcolor[HTML]{C0C0C0}                                                    & DEUP                                    & 50.59                                       & 34.67   &28.23&19.32                                   & \textcolor{red}{10.71}                                   & 14.82                                      & 13.86                                          & 15.55            &5.46                               \\ 
\cellcolor[HTML]{C0C0C0}                                                    & DataSUITE                              & 42.92                                       & 71.06        &37.11&47.6                              & 21.96                                            & 15.26                                      & 14.8                                           & 30.78                              & 12.8             \\ 
\cellcolor[HTML]{C0C0C0}                                                    & \name ~($\mathtt{Score_1}$)                          & {\textcolor{blue} {14.05}}                                 & {\textcolor{blue}{13.62}}     & \textcolor{blue}{11.8}                              & \textcolor{blue}{7.01}                                                        & 12.91                                            & {\textcolor{blue}{12.44}}                                & {\textcolor{blue}{13.33}}                                    & {\textcolor{blue}{12.90}}                       &  \textcolor{red}{2.56}                                             \\ 
\multirow{-4}{*}{\cellcolor[HTML]{C0C0C0}{$\text{C}_{\text{low}}$}$\downarrow$}  & \name ~($\mathtt{Score_2}$)                          & \textcolor{red}{10.13}                              & \textcolor{red}{10.41}    & \textcolor{red}{9.93}                              & \textcolor{red}{6.15}                                                      & 10.93                                            & \textcolor{red}{8.71}                              & \textcolor{red}{10.42}                                 & \textcolor{red}{11.18}                            &  \textcolor{blue}{2.83}                                            \\ \hline \hline
\cellcolor[HTML]{C0C0C0}                                                    & DEUP                                    & 15.47                                       & 12.42          &17.99&7.71                            & 11.28                                            & \textcolor{red}{6.18}                              & 3.36                                           & 23.94                                       & 10.05    \\ 
\cellcolor[HTML]{C0C0C0}                                                    & DataSUITE                              & 37.51                                       & 36.55                 &14.85&6.82                      & \textcolor{red}{5.97}                                    & 10.57                                      & 22.56                                          & 4.23                                     & 18.93        \\ 
\cellcolor[HTML]{C0C0C0}                                                    & \name ~($\mathtt{Score_1}$)                          & \textcolor{red}{8.89}                               & {\textcolor{blue}{10.39}}     & \textcolor{blue}{4.72}                                & \textcolor{blue}{4.12}                                                        & 7.71                                             & 8.09                                       & {\textcolor{blue}{3.19}}                                     & {\textcolor{blue} {1.69}}                     &  \textcolor{blue}{5.22}                 \\ 
\multirow{-4}{*}{\cellcolor[HTML]{C0C0C0}{$\text{C}_{\text{high}}$}$\downarrow$} & \name ~($\mathtt{Score_2}$)                          & {\textcolor{blue}{11.03}}                                 & \textcolor{red}{9.37}     & \textcolor{red}{3.90}                                & \textcolor{red}{2.83}                                                      & {\textcolor{blue}{7.01}}                                       & {\textcolor{blue}{7.30}}                                  & \textcolor{red}{2.95}                                  & \textcolor{red}{1.65}                            &  \textcolor{red}{4.19}           \\ \hline
\end{tabular}
}
\label{tab:HD_gaps}
\end{table*}

\begin{table*}[t]
\centering
\caption{\textbf{Assessing the identified risk regimes for regression benchmarks (Tails) with dimensionality ranging between $2$ and $32, 000$.}  For every metric, we identify the \textcolor{red}{first} and \textcolor{blue}{second} best approach across the different benchmarks.}
\label{tab:HD_tails}
\renewcommand{\arraystretch}{1.15}
\resizebox{0.9\textwidth}{!}{
\begin{tabular}{|
>{\columncolor[HTML]{C0C0C0}}c |
>{\columncolor[HTML]{EFEFEF}}c |c|c|c|c|c|c|c|c|c|}
\hline
                                                          &\cellcolor[HTML]{C0C0C0}&\cellcolor[HTML]{C0C0C0}&\cellcolor[HTML]{C0C0C0}&\cellcolor[HTML]{C0C0C0}&\cellcolor[HTML]{C0C0C0}&\cellcolor[HTML]{C0C0C0}&\cellcolor[HTML]{C0C0C0}&\cellcolor[HTML]{C0C0C0}&\cellcolor[HTML]{C0C0C0}&\cellcolor[HTML]{C0C0C0} \\
\multirow{-2}{*}{\textbf{Metrics}}  & \multirow{-2}{*}{\cellcolor[HTML]{C0C0C0}\textbf{Method}} & \multirow{-2}{*}{ \cellcolor[HTML]{C0C0C0} \textbf{Camel}} & \multirow{-2}{*}{ \cellcolor[HTML]{C0C0C0}\textbf{Levy}} & \multirow{-2}{*}{ \cellcolor[HTML]{C0C0C0}\textbf{Airfoil}} & \multirow{-2}{*}{ \cellcolor[HTML]{C0C0C0}\textbf{NO2}} & \multirow{-2}{*}{ \cellcolor[HTML]{C0C0C0}\textbf{Kinematics}} & \multirow{-2}{*}{ \cellcolor[HTML]{C0C0C0}\textbf{Puma}} & \multirow{-2}{*}{ \cellcolor[HTML]{C0C0C0}\textbf{Housing}} & \multirow{-2}{*}{ \cellcolor[HTML]{C0C0C0}\textbf{Ailerons}} & \multirow{-2}{*}{ \cellcolor[HTML]{C0C0C0}\textbf{DTI}}
                                                          
                                                          \\ \hline \hline
\cellcolor[HTML]{C0C0C0}                                           & DEUP                                    & 10.53                                       & 7.34     &11.28&       13.76                           & 14.39                                             & 16.82                                      & \textcolor{blue}{2.11}                                  & 18.37                                      &  19.23    \\ 
\cellcolor[HTML]{C0C0C0}                                           & Data SUITE                              & 3.84                                        & 9.21    & 11.02&    12.16                                & 17.59                                             & 22.38                                      & 17.89                                          & \textcolor{blue}{17.58}                     &   20.06           \\ 
\cellcolor[HTML]{C0C0C0}                                           & \name ~($\mathtt{Score_1}$)                          & \textcolor{red}{0.0}                                  & \textcolor{red}{4.56}       &\textcolor{red}{1.94}& \textcolor{blue}{3.65}                       & \textcolor{blue}{8.02}                                       & \textcolor{red}{8.78}                               & \textcolor{red}{1.05}                                  & \textcolor{red}{9.59}                              & \textcolor{red}{6.67}       \\ 
\multirow{-4}{*}{\cellcolor[HTML]{C0C0C0}FN$\downarrow$}                       & \name ~($\mathtt{Score_2}$)                          & \textcolor{blue}{0.25}                               & \textcolor{blue}{4.82}     &\textcolor{blue}{2.48}&     \textcolor{red}{3.25}                      & \textcolor{red}{7.18}                                     & \textcolor{blue}{10.38}                              & 2.32                                           & \textcolor{red}{9.59}                            &\textcolor{blue}{7.13}
\\ \hline \hline
\cellcolor[HTML]{C0C0C0}                                           & DEUP                                    & 9.53                                        & 7.35       & 10.82&    9.11                             & 13.02                                               & 14.67                                      & 8.77                                           & 12.01                                            &  17.34  \\ 
\cellcolor[HTML]{C0C0C0}                                           & Data SUITE                              & 3.83                                        & 6.38    & 9.15&     9.75                               & 24.0                                               & 26.67                                      & 19.3                                           & 12.0                                   &   14.07        \\ 
\cellcolor[HTML]{C0C0C0}                                           & \name ~($\mathtt{Score_1}$)                          & \textcolor{red}{0.42}                               & \textcolor{red}{1.68}       & \textcolor{red}{2.85}&  \textcolor{red}{4.27}                          & \textcolor{blue}{6.33}                                    & \textcolor{blue}{13.33}                             & \textcolor{red}{3.51}                                  & \textcolor{blue}{0.80}                              & \textcolor{blue}{9.09}       \\ 
\multirow{-4}{*}{\cellcolor[HTML]{C0C0C0}FP$\downarrow$}                       & \name ~($\mathtt{Score_2}$)                          & \textcolor{blue}{1.68}                               & \textcolor{blue}{2.52}     & \textcolor{blue}{4.29}&  \textcolor{blue}{6.18}                        & \textcolor{red}{6.18}                                    & \textcolor{red}{12.23}                              & \textcolor{blue}{4.26}                                  & \textcolor{red}{0.38}                            & \textcolor{red}{8.36}        \\ \hline \hline
\cellcolor[HTML]{C0C0C0}                                           & DEUP                                    & 34.04                                       & 52.74                             &29.11& 16.95       & \textcolor{blue}{6.36}                                    & \textcolor{blue}{5.37}                              & 13.0                                             & \textcolor{red}{11.07}                           & 48.25   \\ 
\cellcolor[HTML]{C0C0C0}                                           & Data SUITE                              & 42.08                                       & 81.06     &57.01&     33.47                             & 7.34                                             & 5.67                                       & 17.73                                          & 16.52                                &      90.11     \\ 
\cellcolor[HTML]{C0C0C0}                                           & \name ~($\mathtt{Score_1}$)                          & \textcolor{blue}{15.59}                              & \textcolor{blue}{26.44}               &\textcolor{red}{8.25}&    \textcolor{blue}{15.09}           & 6.58                                             & \textcolor{red}{4.61}                              & \textcolor{red}{5.14}                                  & 17.19                                     &   \textcolor{blue}{19.94}   \\ 
\multirow{-4}{*}{\cellcolor[HTML]{C0C0C0}$\text{C}_{\text{low}}$$\downarrow$}  & \name ~($\mathtt{Score_2}$)                          & \textcolor{red}{14.37}                              & \textcolor{red}{14.04}      &\textcolor{blue}{10.08} & \textcolor{red}{11.73}                       & \textcolor{red}{5.73}                                    & 5.5                                        & \textcolor{blue}{6.67}                                  & \textcolor{blue}{11.38}                              & \textcolor{red}{17.01}   \\ \hline \hline
\cellcolor[HTML]{C0C0C0}                                           & DEUP                                    & 23.69                                       & 20.75     & 14.56&   27.34                               & \textcolor{red}{6.83}                                    & \textcolor{blue}{\textcolor{red}{2.63}}                     & 5.69                                           & 7.25                                      &   39.94   \\ 
\cellcolor[HTML]{C0C0C0}                                           & Data SUITE                              & 17.49                                       & 27.32   & 18.09&        31.58                            & 10.08                                            & 6.41                                       & 5.15                                           & 4.97                                        &  64.48  \\ 
\cellcolor[HTML]{C0C0C0}                                           & \name ~($\mathtt{Score_1}$)                          & \textcolor{blue}{7.5}                                & \textcolor{blue}{17.93}                     &\textcolor{blue}{15.19}&   \textcolor{blue}{12.08}      & 7.14                                             & \textcolor{red}{2.46}                              & \textcolor{blue}{5.07}                                  & \textcolor{red}{2.31}                              &  \textcolor{blue}{13.35}   \\ 
\multirow{-4}{*}{\cellcolor[HTML]{C0C0C0}$\text{C}_{\text{high}}$$\downarrow$} & \name ~($\mathtt{Score_2}$)                          & \textcolor{red}{6.7}                                & \textcolor{red}{15.18}       &\textcolor{red}{16.64} &  \textcolor{red}{10.68}                      & \textcolor{blue}{7.09}                                    & 2.81                                       & \textcolor{red}{4.05}                                  & \textcolor{blue}{2.43}                           &   \textcolor{red}{11.06}     \\ \hline
\end{tabular}
}
\end{table*}

\subsection{\name~Framework}
Since it is challenging to accurately estimate and interpret sample-level error estimates, particularly in OOS or OOD regimes, a more tractable approach is to analyze sample groups that correspond to varying levels of expected risk. To this end, \name~organizes a set of test samples into different risk regimes. Without loss of generality, we assume that a typical test set contains samples close to the training distribution, as well as OOS and potential OOD samples.

In our implementation, both scores are split into three bins using conditional quantile ranges (\textit{low}:$[0,25]$, \textit{moderate}:$[25,75]$ and \textit{high}:$[75,100]$), thereby creating a non-trivial partition of the test data into risk regimes. Note that, the number of bins and the threshold choices used are only for demonstration, and can be adapted based on application needs (e.g., pick top $k\%$ of high-risk samples). As discussed earlier, \name~does not involve any calibration step and can directly work on the unlabeled test set. We now describe the different risk regimes in \name.

\noindent \textbf{ID} (\textcolor[HTML]{48893D} {$\blacksquare$}): The model generalizes well in this regime and is expected to produce low prediction error. In \name, this corresponds to samples with low uncertainty as well as low/moderate MNC scores;

\noindent \textbf{Low Risk} (\textcolor[HTML]{B57E2A} {$\blacksquare$}): Even when the uncertainty is low, the model can produce higher error than the ID samples, when there is incongruity (e.g., samples within a neighborhood having different target values). Similarly, for OOS samples with moderate uncertainties, the model can still extrapolate well and produce reduced risk. Hence, we define this regime as the collection of (low uncertainty, high MNC) and (moderate uncertainty, low/moderate MNC) samples;

\noindent \textbf{Moderate Risk} (\textcolor[HTML]{6075F6} {$\blacksquare$}): Since epistemic uncertainties can be inherently miscalibrated, OOS samples, which the model cannot extrapolate to, can be associated with moderate uncertainties. On the other hand, the model could reasonably generalize to OOD samples that are flagged with high uncertainties. Hence, we define this regime as the collection of (moderate uncertainty, high MNC) and (high uncertainty, low/moderate MNC) samples;

\noindent \textbf{High Risk} (\textcolor[HTML]{992F28} {$\blacksquare$}): Finally, when both the uncertainty and non-conformity scores are high, there is no evidence that the model will behave predictably on those samples. In practice, this can correspond to both OOS and OOD samples.

\subsection{Evaluation Metrics}
Evaluation metrics typically adopted to assess failure detectors for regression models, e.g., Spearman correlation between the true risk and the predicted risk
on a held-out test set or the average error in top inconsistent samples, do not comprehensively reflect the quality of detectors in different risk regimes. Hence, we propose to utilize the following metrics (see Figure \ref{fig:framework}):

\noindent \textbf{False Negatives (FN)($\downarrow$)} This is the most important metric in applications where the cost of missing to detect high risk failures is high. We measure the ratio of samples in the ID or low risk regimes that actually have high true risk (top $20^{\text{th}}$ percentile of all test samples).

\noindent \textbf{False Positives (FP)($\downarrow$)} This reflects the penalty for scenarios where arbitrarily flagging harmless samples as failures. Here, we measure the ratio of samples in the moderate or high risk regimes that actually have low true risk (bottom $20^{\text{th}}$ percentile of all test samples).

\noindent \textbf{Confusion in Low Risk Regimes (C$_{\text{low}}$)($\downarrow$)} A common challenge in fine-grained sample grouping (ID vs low risk) is that the detector can confuse samples between neighboring regimes. We define this metric as the ratio between the $90^{\text{th}}$ percentile of the ID regime and the $10^{\text{th}}$ percentile of the low risk regime. The selection of the $90^{\text{th}}$ percentile and $10^{\text{th}}$ percentile to gauge the error ratio is intentionally made stringent. However, one can relax these thresholds depending on the desired error tolerance in practice.

\noindent \textbf{Confusion in High Risk Regimes (C$_{\text{high}}$)($\downarrow$)} This is similar to the previous case and instead measures the confusion between the moderate and high risk regimes.

\section{Experiments}
\label{sec:setup}

\textbf{Datasets.} We evaluate the effectiveness of \name~using a suite of tabular and imaging benchmarks.  
\begin{enumerate}[leftmargin=*]
\item \underline{1D Benchmark Functions}: We consider the following black-box functions t:
\begin{enumerate}
    \item $f_1(\mathrm{x})$ = $\begin{cases*}
                    \phantom{-}\mathrm{x}^2 & if  $\mathrm{x} < 2.25 \phantom{x} \text{or} \phantom{x} \mathrm{x} > 3.01$   \\
                     \phantom{-}\mathrm{x}^2 - 20 & otherwise
                 \end{cases*}$ (Figure 1)
    \item $f_2(\mathrm{x}) = \sin(2\pi \mathrm{x})$, $\mathrm{x} \in [-0.5, 2.5]$ 
    \item $f_3(\mathrm{x}) = a \exp(-b \mathrm{x}) + \exp(\cos(c \mathrm{x})) - a - \exp(1)$, $\mathrm{x} \in [-5, 5]$, $a = 20, b= 0.2, c= 2\pi$
    \item $f_4(\mathrm{x}) = \sin(\mathrm{x})\cos(5\mathrm{x})\cos(22\mathrm{x})$, $\mathrm{x} \in [-1, 2]$
\end{enumerate} In each of these functions, we used $200$ test samples drawn from an uniform grid and computed the metrics.

\item \underline{HD Regression Benchmarks}: We also considered a set of regression datasets comprising different domains and varying dimensionality. (a) Camel (2D), (b) Levy (2D)~\citep{bench_fun} characterized by multiple local minima, (c) Airfoil (5D), (d) NO2 (7D), (e) Kinematics (8D), (f) Puma (8D)~\citep{delve} which are simulated datasets of the forward dynamics of different robotic control arms, (g) Boston Housing (13D)~\citep{bh}, (h) Ailerons (39D)~\citep{ailerons} which is a dataset for predicting control action of the ailerons of an F16 aircraft, and (i) Drug-Target Interactions (32000D). For each benchmark, we created two variants: Gaps (training exposed to data with targets between $(0-30^{th})$ and $(60-100^{th})$ percentiles) and Tails (training exposed to $(0-70^{th})$ percentiles of the targets). Additionally, we considered the Skillcraft dataset~\citep{cmixup}, which represents real-world distribution shifts arising from change in the league index. 
\item \underline{Image Regression:} We used three image regression benchmarks namely chair (yaw) angle, cell count and CIFAR-10 rotation prediction respectively. In each case, we synthesized two different variants -- tails and gaps in the target variable, similar to the HD regression experiments. The range of target values used in each of the experiments can be found in Figure \ref{fig:image_regimes}. 
\end{enumerate}

\textbf{Baselines}. \underline{(i) DEUP}~\citep{deup} is a state-of-the-art epistemic uncertainty-based failure detection approach. It utilizes a post-hoc, auxiliary error predictor that learns to predict the risk of the underlying model which is considered as a surrogate for uncertainties; \underline{(ii) DataSUITE}~\citep{datasuite} is a task-agnostic approach that estimates the inconsistencies in the data regimes to assess data quality. Both baselines rely on the use of additional, curated calibration data to either train the error predictor in case of DEUP, or to obtain non-conformity scores that assess the sample level quality in the latter. 

\textbf{Training Protocols}. For experiments on all tabular benchmarks, we used an MLP~\citep{bishop} with $4$ layers each with a hidden dimension of $128$. While we used the WideResNet40-2 model~\citep{zagoruyko2016wide} for the first two image regression datasets, in the case of CIFAR-10, we randomly applied a rotation transformation [0 - 90 degrees] to each 32$\times$32$\times$3 image and trained a ResNet-34 model to predict the angle of rotation. For evaluation, we used the held-out test sets (e.g., 10K randomly rotated images for CIFAR-10). Without loss of generality, we used the $L_{1}$ objective for training all the models. We provide the implementation details along with hyper-parameters choices in Appendix \ref{app: 3}.

\section{Main Findings \& Discussion}
\label{sec:results}

\subsection{Results on 1D benchmarks} We expect an effective failure detector to align well with the training distribution (ID) and progressively flag regions of low, moderate and high risk as we move away from the inferred data manifold. From the results in Table \ref{tab:1d} for standard $1$D benchmark functions, \name~achieves this effectively, while also consistently outperforming the baselines across all the metrics. Furthermore, from Figure \ref{fig:teaser}, we notice that \name~accurately identifies the training data regimes (\textcolor[HTML]{48893D}{Green}) as ID. As we traverse further from the training manifold, \name~assigns low risk (\textcolor[HTML]{B57E2A}{Yellow}) to unseen examples that are close to the training data. Notably, as we encounter samples that are significantly out-of-distribution, it consistently flags them as moderate/high risk. As an ablation, we also include the performance obtained by (a) using only the MNC ($\mathtt{Score}_1$) and (b) using only uncertainties from \name, in order to demonstrate the importance of considering them jointly. While the MNC-only baseline can reasonably control FP, it is not able to reduce the FN. Since those scores are unnormalized, they behave differently across different data regimes. Hence, they are insufficient to accurately rank samples on their own. On the other hand, we observe that Anchor-UQ is a stronger baseline, even outperforming DEUP in many cases.

In addition to the fidelity metrics, computational efficiency is another important aspect of failure detectors. Hence, we provide the inference run-times for each of the methods, measured using a test set of 1000 samples on the 1D benchmarks with a single GPU. While DataSUITE (\textbf{29.8s}) involves training an autoencoder followed by conformalization, DEUP (\textbf{18.2s}) requires training an auxiliary risk estimator to evaluate risk. In comparison, computing $\mathtt{Score_1}$ with \name~is very efficient (\textbf{1.55s}) as it basically involves only forward passes with the anchored model. While $\mathtt{Score_2}$ comes with an increased computational cost (40.9s), we find that it helps in resolving regimes of moderate and high risk better, and handling corruptions at test time (Figure \ref{fig:corruption}).

\begin{figure}[t]
    \centering
    \includegraphics[width=0.7\columnwidth]{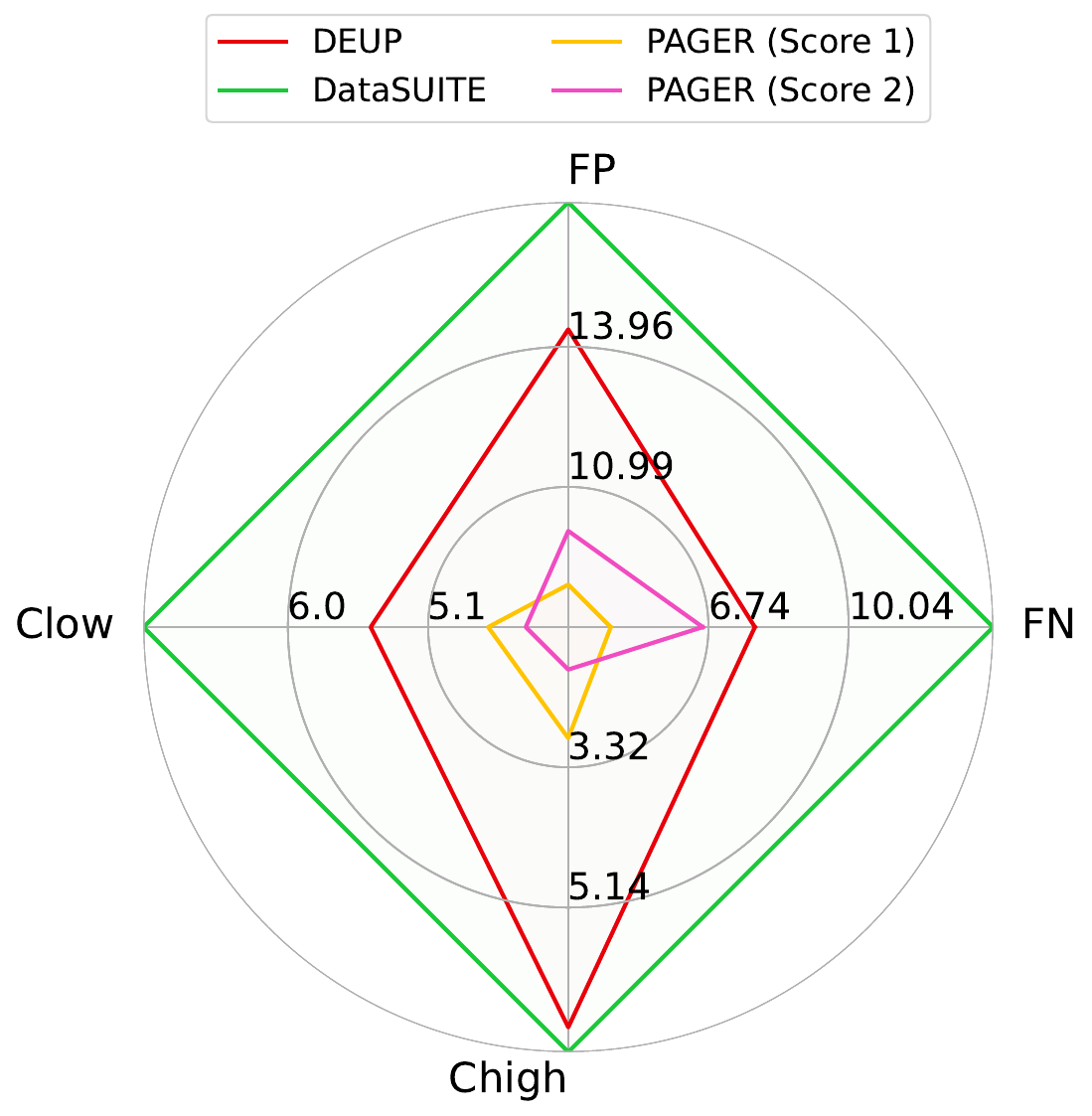}
    \caption{\textbf{\name~can detect failures under complex distribution shifts effectively.} We assess \name~on the Skillcraft dataset characterized by real-world shifts (change in league index), and find that it achieves reductions in all metrics over the baselines.}
    \label{fig:skillcraft}
\end{figure}

\begin{figure*}[t]
    \centering
    \includegraphics[width=1\linewidth]{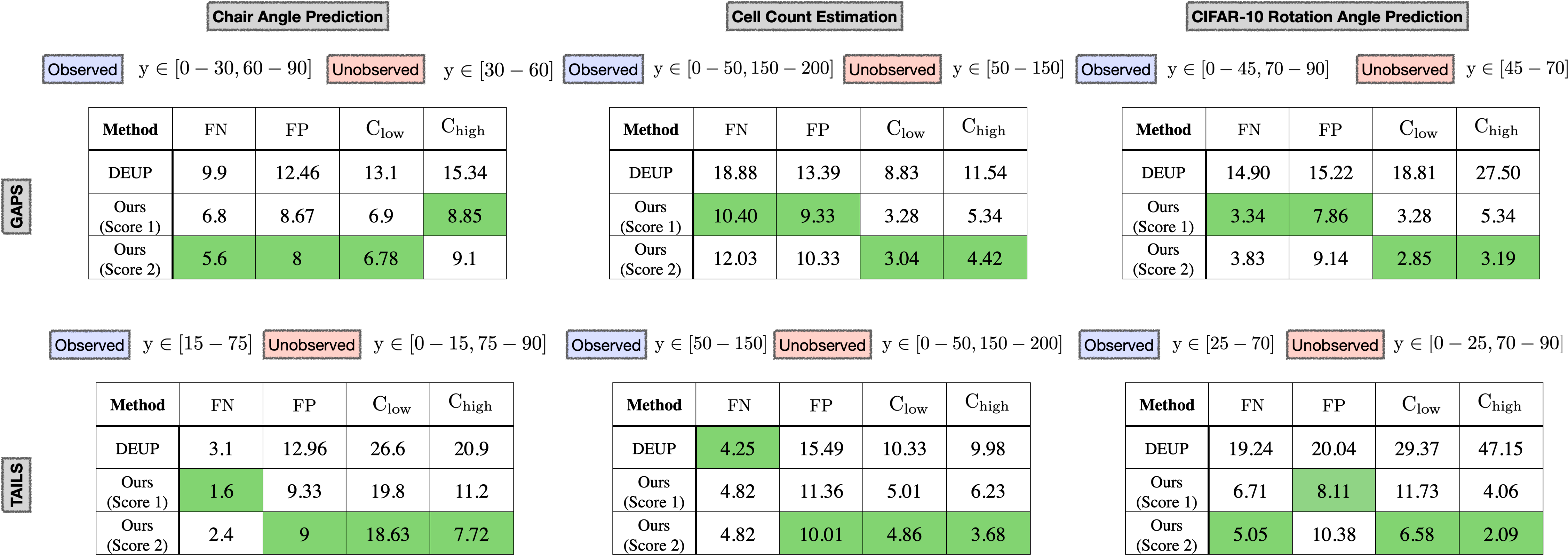}
    \caption{\textbf{Efficacy of \name~on Image Regression Benchmarks.} We can observe that in comparison to the state-of-the art baseline DEUP, \name~effectively minimizes the FN, FP and confusion metrics even under challenging extrapolation scenarios. We find that \name~can consistently flag samples from the unobserved regimes which corresponds to highly erroneous predictions.}
    \label{fig:image_regimes}
\end{figure*}

\subsection{Results on HD Regression Datasets} The observations from our 1D experiments persist even with higher dimensional regression benchmarks, thus evidencing the efficacy of \name. From Tables \ref{tab:HD_gaps} and \ref{tab:HD_tails}, we find that even in higher dimensions and more complex extrapolation scenarios (e.g., Gaps and Tails, as discussed in Section~\ref{sec:setup}), \name~is able to produce $> 50\%$ reduction in FN and FP metrics over the baselines. Furthermore, \name~significantly reduces the amount of overlap (C$_{\text{low}}$ and C$_{\text{high}}$) between the risk regimes. The baselines on the other hand produce higher confusion scores demonstrating their limitations in risk stratification. This observation persists even on the Skillcraft dataset containing real-world distribution shifts (Figure \ref{fig:skillcraft}). Finally, despite the increased computational complexity, $\mathtt{Score_2}$ leads to lower confusion scores compared to $\mathtt{Score_1}$ while producing comparable FP and FN metrics.

\subsection{Results on Imaging Benchmarks} Our analysis in Figure \ref{fig:image_regimes} reveals that \name~achieves lower FN, FP, and confusion scores compared to the baselines, even when confronted with challenging extrapolation regimes in imaging datasets. This demonstrates the effectiveness of our approach in handling diverse modalities of data. Additionally, we provide sample images that were accurately identified as high risk by \name~in Appendix \ref{app: 5}. Notably, these examples correspond to regimes that were not encountered during training. 

\begin{table}[t]
\centering
\caption{\textbf{Predictive performance of anchoring.} The use of anchored training for failure characterization does not compromise on the performance ($R^{\text{2}}$ scores), regardless of whether the test data comes from  observed or unobserved regimes.}
\renewcommand{\arraystretch}{1.15}
\vspace{0.1in}
\resizebox{0.99\columnwidth}{!}{
\begin{tabular}{|c|c|c|c|}
\hline
\cellcolor[HTML]{C0C0C0}{\textbf{Dataset}}&\cellcolor[HTML]{C0C0C0}{\textbf{Training}}&\cellcolor[HTML]{C0C0C0}{\textbf{$R^{\text{2}}$ (Observed)}}&\cellcolor[HTML]{C0C0C0}{\textbf{$R^{\text{2}}$ (Unobserved)}}
\\ \hline \hline
CIFAR-10& Standard & $0.92$ & {$0.77$} \\
(rotation angle)&Anchoring&\textcolor{red}{$0.93$}&\textcolor{red}{$0.81$} \\
\hline
Chairs& Standard & \textcolor{red}{$0.97$} & $0.73$ \\
(yaw angle)&Anchoring&\textcolor{red}{$0.97$}&\textcolor{red}{$0.75$} \\
\hline
Cells& Standard & $0.88$ & {$0.69$} \\
(count)&Anchoring&\textcolor{red}{$0.89$}&\textcolor{red}{$0.72$} \\
\hline
\end{tabular}
}
\label{tab:perf}
\end{table}

\begin{figure}[t]
    \centering
    \includegraphics[width=0.9\columnwidth]{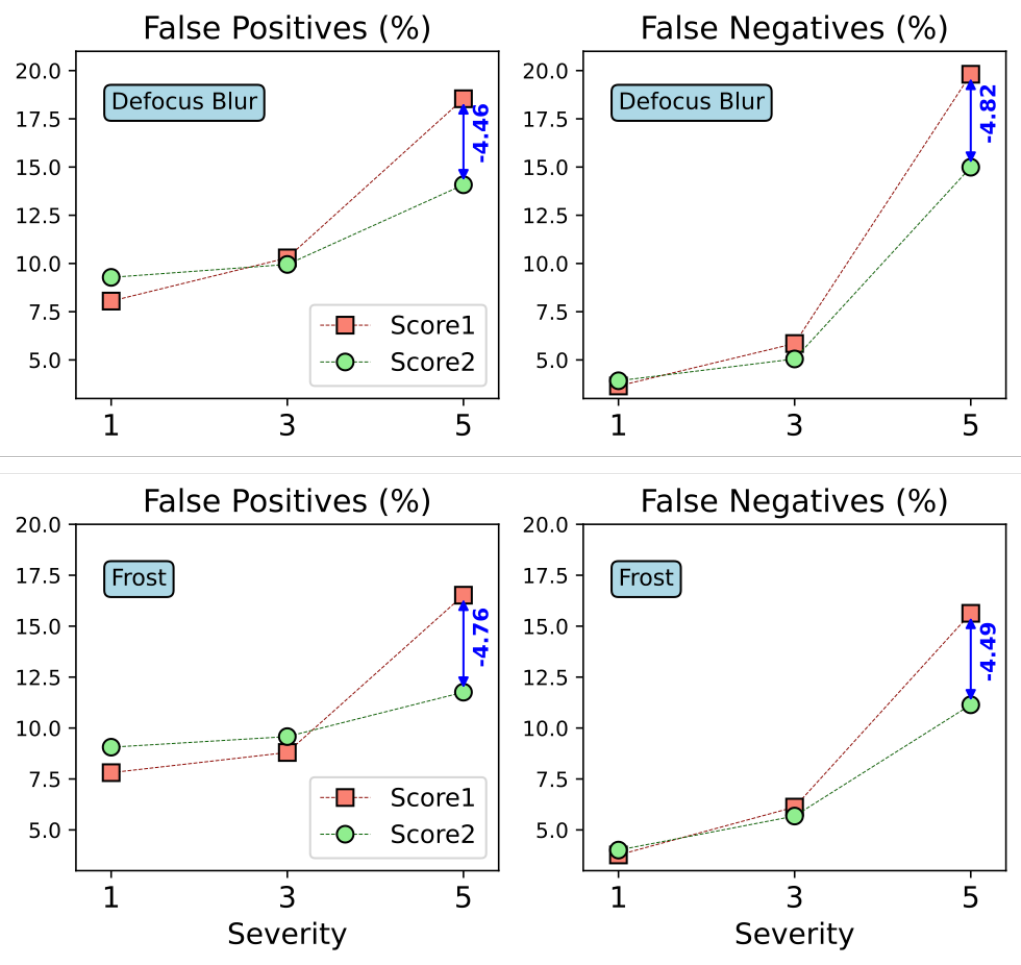}
    \caption{\textbf{Benefits of $\mathtt{Score_2}$.} Under test time image corruptions such as defocus blur and frost, $\mathtt{Score_2}$ can better resolve these risk regimes and reduce both FP and FN metrics over $\mathtt{Score_1}$.}
    \label{fig:corruption}
\end{figure}

\subsection{Analysis}
In this section, we provide deeper empirical insights into the behavior of the proposed approach.

\textbf{A. Do anchored models compromise performance?} An important question that pertains the use of anchored models for failure characterization is whether these models compromise predictive performance. Anchored training is a general protocol applicable to any architecture and typically leads to improved generalization, particularly under distribution shifts. Conceptually, as showed in~\cite{thiagarajan2022single}, centering a dataset using different constant inputs will lead to different solutions, due to inherent lack of shift invariance in neural tangent kernels induced by commonly adopted neural networks. Building upon this principle, we use different anchors for the same sample across different epochs with the goal of marginalizing out the effect of anchor choice at inference time. Through this process, anchoring implicitly enables the training process to explore a richer class of hypotheses, and often produces improved predictive performance when compared to standard model training. To demonstrate this, we computed the test performance ($R^2$ statistic) in both observed (range of $y$ values exposed during training) and unobserved (range of $y$ values unseen during training) regimes for the three image regression benchmarks. As shown in Table \ref{tab:perf}, anchoring performs competitively and sometimes even outperforms standard training.

\textbf{B. When should one use $\mathtt{Score_2}$ in PAGER?} As shown above, $\mathtt{Score_2}$ often demonstrates noteworthy improvements in confusion scores ($\text{C}_{\text{low}}$ and $\text{C}_{\text{high}}$). This is valuable in scenarios where users need to flag samples with the highest risk, and ensure that high-risk samples are not misclassified as moderate risk. Another scenario where $\mathtt{Score_2}$ is beneficial is when the test samples are drawn from a different distribution compared to training (referred to as covariate shifts). To demonstrate this, we repeated the CIFAR-10 rotation angle prediction experiment by applying natural image corruptions (defocus blur and frost) at varying severity levels (Figure \ref{fig:corruption}). Interestingly, we observed significant improvements in both FP and FN scores with $\mathtt{Score_2}$ as the severity increased. In summary, while $\mathtt{Score_1}$ excels in scalability and is well-suited for online evaluation, $\mathtt{Score_2}$ effectively addresses challenging testing scenarios.

\begin{figure}[t]
    \centering
    \includegraphics[width=0.9\columnwidth]{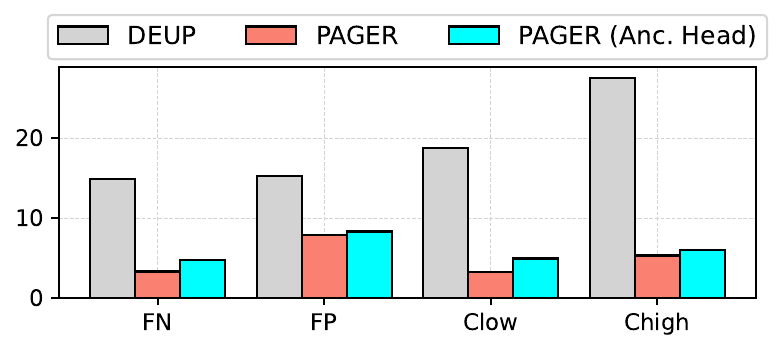}
    \caption{\textbf{\name~can be implemented for a model trained without anchoring.} By training only a regression head via anchoring, one can implement \name~with a pre-trained feature extractor backbone. Using the CIFAR-10 rotation angle prediction experiment, we show that this variant produces improvements over the DEUP baseline, similar to \name~with a fully anchored model.}
    \label{fig:head}
\end{figure}

\begin{table}[t]
\centering
\caption{\textbf{Impact of post-hoc calibration on PAGER.} When additional calibration data becomes available, PAGER can leverage it to both recalibrate the uncertainty estimates as well as improve the non-conformity scores. This leads to consistently superior metrics over the PAGER variant without additional calibration data.}
\renewcommand{\arraystretch}{1.15}
\vspace{0.1in}
\resizebox{0.99\columnwidth}{!}{
\begin{tabular}{|c|c|c|c|c|c|c|}
\hline
\cellcolor[HTML]{C0C0C0}{\textbf{Dataset}}&\cellcolor[HTML]{C0C0C0}{\textbf{Method}}&\cellcolor[HTML]{C0C0C0}{\textbf{Calibration?}}&\cellcolor[HTML]{C0C0C0}{\textbf{FP}$\downarrow$}&
\cellcolor[HTML]{C0C0C0}{\textbf{FN}$\downarrow$}&
\cellcolor[HTML]{C0C0C0}{\textbf{$\text{C}_{\text{low}}$}$\downarrow$}&
\cellcolor[HTML]{C0C0C0}{\textbf{$\text{C}_{\text{high}}$}$\downarrow$}
\\ \hline \hline

& & No &$17.34$&$19.23$&$48.25$&$39.94$\\
& \multirow{-2}{*}{DEUP} &Yes &$13.08$&$12.95$&$33.10$&$19.26$\\
\cline{2-7}
& & No &$9.09$&$6.67$&$19.94$&$13.35$\\
\multirow{-4}{*}{DTI}&\multirow{-2}{*}{PAGER} & Yes &\textcolor{red}{$4.27$}&\textcolor{red}{$3.88$}&\textcolor{red}{$17.55$}&\textcolor{red}{$10.76$}
\\
\hline
& & No &$15.22$&$14.90$&$18.81$&$27.50$\\
& \multirow{-2}{*}{DEUP} &Yes &$6.08$&$4.64$&$6.36$&$7.75$\\
\cline{2-7}
& & No &$7.86$&$3.34$&$3.28$&$5.34$\\
\multirow{-4}{*}{CIFAR-10}&\multirow{-2}{*}{PAGER} & Yes &\textcolor{red}{$4.81$}&\textcolor{red}{$2.59$}&\textcolor{red}{$1.70$}&\textcolor{red}{$4.48$}
\\
\hline
\end{tabular}
}
\label{tab:posthoc}
\end{table}

\textbf{C. Implementing PAGER with an anchored regression head.} Regarding the application of \name~to models trained without anchoring, we first emphasize that anchoring does not necessitate any adjustments to the optimizer, loss function, or training protocols. However, in situations involving pre-trained models, it is feasible to train solely an anchoring-based regression head attached to an existing feature extractor. Optionally, one may fine-tune the feature extractor concurrently with the anchored regression head in an end-to-end manner, adhering to standard practices. As an illustration, in the experiment on CIFAR-10 rotation angle prediction under the Gaps setting with $\mathtt{Score_1}$, we considered a variant, where we trained an anchored head while keeping the feature extractor frozen. Note, the feature extractor was obtained through standard training on the same dataset. Our findings in Figure \ref{fig:head} reveal that even with this approach, the performance is comparable to \name~implemented with a fully anchored model.

\textbf{D. Does post-hoc calibration help PAGER?}
In real-world applications, obtaining access to calibration data representing the distribution shifts is not always practical. Hence, one of our objectives was to ensure that that PAGER can still meaningfully organize samples into different risk regimes, even without post-hoc calibration. However, it is common practice to adopt post-hoc calibration, when additional data becomes available. In such scenarios, PAGER can leverage the data to (i) recalibrate the uncertainty estimates from forward anchoring for guaranteed coverage, and (ii) improve the quality of the non-conformity score estimates. To demonstrate this, we performed an additional experiment, where we calibrated the uncertainty estimates (90\% coverage), and reimplemented the non-conformity score computation by obtaining the max discrepancy over the union of training ($\mathcal{D}$) and calibration sets ($\mathcal{D}_c$). From Table \ref{tab:posthoc}, we find that the performance of both the baseline and PAGER improve by including additional calibration data, and more importantly, the benefits of PAGER persist.

\begin{table}[t]
\centering
\caption{\textbf{Ablation study.} Using only uncertainties or the proposed non-conformity scores leads to sub-par performance in risk characterization on the CIFAR-10 benchmark.}
\renewcommand{\arraystretch}{1.15}
\vspace{0.1in}
\resizebox{0.72\columnwidth}{!}{
\begin{tabular}{|c|c|c|c|c|}
\hline
\cellcolor[HTML]{C0C0C0}{\textbf{Method}}&\cellcolor[HTML]{C0C0C0}{\textbf{FP}$\downarrow$}&
\cellcolor[HTML]{C0C0C0}{\textbf{FN}$\downarrow$}&
\cellcolor[HTML]{C0C0C0}{\textbf{$\text{C}_{\text{low}}$}$\downarrow$}&
\cellcolor[HTML]{C0C0C0}{\textbf{$\text{C}_{\text{high}}$}$\downarrow$}
\\ \hline \hline
UQ-only & $12.54$ &$13.45$ &$14.75$& $9.15$\\
MNC-only & $13.08$ &$13.24$ &$12.91$& $11.38$\\
PAGER & \textcolor{red}{$5.05$} &\textcolor{red}{$10.38$} &\textcolor{red}{$6.58$}& \textcolor{red}{$2.09$}\\
\hline

\end{tabular}
}
\label{tab:ablation}
\end{table}

\textbf{E. Ablation study.} In order demonstrate the value of combining uncertainty and manifold non-conformity scores in \name, we performed an ablation study on the CIFAR-10 rotation angle benchmark. In theory, all extreme rotations should be picked by large uncertainties -- however, in practice, they tend to produce both false positives and false negatives, which can be attributed to insufficiency of epistemic uncertainties (see Figure \ref{fig:teaser}) as well as their poor calibration under distribution shifts. On the other hand, while non-conformity can identify discrepancies arising due to feature inconsistency, it has the inherent challenge that it only measures the relative change in the target space or distances in the input space. Since these scores are unnormalized, they can vary vastly across different uncertainty regimes. Consequently, as we illustrate in Table \ref{tab:ablation}, combining both scores leads to significantly improved performance.

\section{Conclusions}
In this paper, we proposed \name, a framework for failure characterization in deep regression models. It leverages the principle of anchoring to integrate epistemic uncertainties and novel non-conformity scores, enabling the organization of samples into different risk regimes and facilitating a comprehensive analysis of model errors. We identify two key impacts. First, \name~can enhance the safety of AI model deployment by preemptively detecting failure cases in real-world applications. This can prevent costly errors and mitigate risks associated with inaccurate predictions. Second, \name~contributes to advancing research in failure characterization for deep regression.

\section*{Impact Statement}
This paper presents work whose goal is to advance the field of Machine Learning. There are many potential societal consequences of our work, none which we feel must be specifically highlighted here.

\section*{Acknowledgements}
This work was performed under the auspices of the U.S. Department of Energy by the Lawrence Livermore National Laboratory under Contract No. DE-AC52-07NA27344. Supported by the LDRD Program under project 22-SI-004. LLNL-CONF-850978.

\bibliography{main}
\bibliographystyle{icml2024}

\newpage
\appendix
\onecolumn
\section{Appendix}
\subsection{Detailed Description of Anchoring in \name}
\label{app: 1}
PAGER expands on the recent successes in anchoring~\cite{thiagarajan2022single, netanyahu2023learning} by building upon the $\Delta$-UQ methodology introduced in~\cite{thiagarajan2022single}. This methodology is used to estimate prediction uncertainties, which play a vital role in characterizing model risk regimes, as depicted in Figure 2 of the main paper. With that context, we now provide a concise overview of $\Delta-$UQ, its training and uncertainty estimation.

\underline{Overview}:~$\Delta-$UQ, short for $\Delta-$Uncertainty Quantification, is a highly efficient strategy for estimating predictive uncertainties that leverages anchoring. It belongs to the category of methods that estimate uncertainties using a single model~\cite{van2020uncertainty, liu2020simple}. $\Delta-$UQ has been demonstrated to be an improved and scalable alternative to Deep Ensembles~\cite{lakshminarayanan2017simple}, eliminating the need to train multiple independent models for estimating uncertainties. The core idea behind $\Delta-$UQ is based on the observation that the injection of constant biases (anchors) to the input dataset produces different model predictions as a function of the bias. To that end, models trained using the same dataset but shifted by respective biases generates diverse predictions. This phenomenon arises from the fact that the neural tangent kernel (NTK)\cite{jacot2018neural} induced in deep models lacks invariance to input data shifts~\cite{bishop}. Consequently, the variance among these models \textit{a.k.a anchored ensembles} serves as a strong indicator of predictive uncertainty. Based on this observation, $\Delta-$UQ follows a simple strategy to consolidate the anchored ensembles into a single model training, where the input is reparameterized as an anchored tuple, as described in Section 2 of the main paper. It is important to note that $\Delta-$UQ performs anchoring in the input space for both vector-valued and image data.

\underline{Training}: In this phase, for every training pair \{$\mathrm{x}, \mathrm{y}$\} drawn from the dataset $\mathcal{D}$, a random anchor $\mathrm{r}_k$ is selected from the same training dataset. Both the input $\mathrm{x}$ and the anchor $\mathrm{r}_k$ are transformed into a tuple given as $[\mathrm{r}_k, \mathrm{x}-\mathrm{r}_k]$. Importantly, this reparameterization does not alter the original predictive task, but instead of using only $\mathrm{x}$, the tuple $[\mathrm{r}_k, \mathrm{x}-\mathrm{r}_k]$ is mapped to the target $\mathrm{y}$. For vector-valued data, $\Delta-$UQ constructs the tuples by concatenating the anchor $\mathrm{r}_k$ and the residual along the dimension axis. In the case of images, the tuples are created by appending along the channel axis, resulting in a $6-$channel tensor for each $3$-channel image. These tuples are organized into batches and used to train the models. Throughout the training process, in expectation, every sample $\mathrm{x}$ is anchored with every other sample in the dataset. The goal here is that the predictions for every $\mathrm{x}$ should remain consistent regardless of the chosen anchor. The training objective is given by:
\begin{equation}
    \theta^* = \arg \underset{\theta}{\min}~~~\mathcal{L}(\mathrm{y}, F_\theta([\mathrm{r}_k, \mathrm{x}-\mathrm{r}_k]),  
\end{equation}
where $\mathcal{L}(.)$ is a loss function such as MAE or MSE. In effect, the $\Delta-$UQ training enforces that
for every input sample $\mathrm{x}$, $F_\theta([\mathrm{r}_1, \mathrm{x}-\mathrm{r}_1]) = F_\theta([\mathrm{r}_2, \mathrm{x}-\mathrm{r}_2]) = \dots = F_\theta([\mathrm{r}_k, \mathrm{x}-\mathrm{r}_k])$, where $F_\theta$ is the underlying model that operates on the tuple $([\mathrm{r}_k, \mathrm{x}-\mathrm{r}_k])$ to predict $\mathrm{y}$.

\underline{Uncertainty Estimation}: During the inference phase, using the trained model with weights $\theta^*$, we compute the prediction $\mathrm{y_t}$ for any test sample $\mathrm{x_t}$. This is performed by averaging the predictions across $K$ randomly selected anchors drawn from the training dataset. The standard deviation of these predictions is then used as the estimate for predictive uncertainty. The equations for calculating the mean prediction and uncertainty around a sample can be found in Equation 1 of the main paper.

\subsection{Algorithm Listings for \name}
\label{app: 2}
Algorithms \ref{algo1},\ref{algo2} and \ref{algo3} provide the details for estimating predictive uncertainty, non-conformity scores - $\mathtt{Score_1}$ and $\mathtt{Score_2}~$ respectively in \name. 

\begin{algorithm}[!t]
\caption{\name: Predictive Uncertainty Estimation}
\label{algo1}
\begin{algorithmic}[1]
\STATE \textbf{Input}:~ Input test samples $\{\mathrm{x}^{t}_i\}_{i=1}^{N}$,~Pre-trained anchored model $F_{\theta^*}$,~ Anchors $\{\mathrm{r}_k\}_{k=1}^{K}$ drawn from the training dataset $\mathcal{D}$  \\
\STATE \textbf{Output}:~ Predictive Uncertainties ($\mathtt{Unc}$) for $\{\mathrm{x}^{t}_i\}_{i=1}^{N}$
\STATE \textbf{Initialize}: $\mathtt{Unc} = \mathtt{list}()$ 
\FOR{$i$ in $1$ to $N$}
\STATE $\mu(y^t_i|\mathrm{x}^t_i) = \frac{1}{K}\sum_{k=1}^K F_{\theta^*}([\mathrm{r}_k,\mathrm{x}^t_i-\mathrm{r}_k] )$; 
\STATE $\sigma(y^t_i|\mathrm{x}^t_i) = \sqrt{\frac{1}{K-1}\sum_{k=1}^K (F_{\theta^*}([\mathrm{r}_k,\mathrm{x}^t_i-\mathrm{r}_k]) - \mu(y^t_i|\mathrm{x}^t_i))^2 }$;
\STATE $\mathtt{Unc[i]} = \sigma(y^t_i|\mathrm{x}^t_i)$
\ENDFOR
\STATE \textbf{return}: $\mathtt{Unc}$
\end{algorithmic}
\end{algorithm}

\begin{algorithm}[!t]
\caption{\name: $\mathtt{Score_1}$ Computation}
\label{algo2}
\begin{algorithmic}[1]
\STATE \textbf{Input}:~ Input test samples $\{\mathrm{x}^{t}_i\}_{i=1}^{N}$,~Pre-trained anchored model $F_{\theta^*}$,~ Train data subset $\{\mathrm{r}_k, \mathrm{y}_k\}_{k=1}^{K}$  \\
\STATE \textbf{Output}:~ $\mathtt{Score_1}$ for $\{\mathrm{x}^{t}_i\}_{i=1}^{N}$
\STATE \textbf{Initialize}: $\mathtt{Score_1} = \mathtt{list}()$ 
\FOR{$i$ in $1$ to $N$}
\STATE $\mathtt{s} = \underset{k}{\max} \bigg\|\mathrm{y}_k - F_{\theta^*}([\mathrm{x}^t_i, \mathrm{r}_k - \mathrm{x}^t_i])\bigg\|_1 ~~\forall k 
 \in \{1 \cdots K\}$;
\STATE $\mathtt{Score_1[i]} = \mathtt{s}$
\ENDFOR
\STATE \textbf{return}: $\mathtt{Score_1}$
\end{algorithmic}
\end{algorithm}

\begin{algorithm}[!t]
\caption{\name: $\mathtt{Score_2}$ Computation}
\label{algo3}
\begin{algorithmic}[1]
\STATE \textbf{Input}:~ Input test samples $\{\mathrm{x}^{t}_i\}_{i=1}^{N}$,~Pre-trained anchored model $F_{\theta^*}$,~Pre-trained anchored auto-encoder $A$,~ Train data subset $\{\mathrm{r}_k, \mathrm{y}_k\}_{k=1}^{K}$, Learning rate $\eta$, Weighing Factor $\lambda$,  No. of iterations $T$\\
\STATE \textbf{Output}:~ $\mathtt{Score_2}$ for $\{\mathrm{x}^{t}_i\}_{i=1}^{N}$
\STATE \textbf{Initialize}: $\mathtt{Score_2} = \mathtt{list()}$ 
\FOR{$i$ in $1$ to $N$}
\STATE \textbf{Initialize}: $\bar{\mathrm{x}} \leftarrow \mathrm{x}^{t}_i$ 
\FOR{$iter$ in $1$ to $T$}
\STATE Compute $\mathcal{R}(\bar{\mathrm{x}}) = \bigg\|\bar{\mathrm{x}} - A([\mathrm{x}^t_i, \bar{\mathrm{x}} - \mathrm{x}^t_i])\bigg\|_2 + \bigg\|\mathrm{x}^t_i - A([\bar{\mathrm{x}}, \mathrm{x}^t_i - \bar{\mathrm{x}}])\bigg\|_2.$
\STATE Compute $L =  \frac{1}{K} \underset{k}{\sum}\|\mathrm{y}_{k} - F_{\theta^*}([\bar{\mathrm{x}}, \mathrm{r}_k - \bar{\mathrm{x}}])\|_1 + \lambda \mathcal{R}(\bar{\mathrm{x}}) $ 
\STATE Update $\bar{\mathrm{x}} \leftarrow \bar{\mathrm{x}} - \eta \nabla_{\bar{\mathrm{x}}}L$
\ENDFOR
\STATE $\mathtt{Score_2[i]} = \|\mathrm{x}^{t}_i - \bar{\mathrm{x}} \|_2$
\ENDFOR
\STATE \textbf{return}: $\mathtt{Score_2}$
\end{algorithmic}
\end{algorithm}

\subsection{Description of our Training Protocols}
\label{app: 3}

In the case of Cell Count and Chair Angle benchmarks, we train an anchored $40-2$ WideResnet model. The training is performed with a batch size of $128$ for $100$ epochs. We utilize the ADAM optimizer with momentum parameters of $(0.9, 0.999)$ and a fixed learning rate of $1e-4$. To train the anchored auto-encoder for computing $\mathtt{Score_2}$, we employ a convolutional architecture with an encoder-decoder structure. The encoder consists of two convolutional layers with kernel sizes of $(3,3)$ and appropriate padding, as well as MaxPooling operations. The decoder comprises two transposed convolutional layers with stride $2$ to reconstruct the input images. We train the anchored auto-encoder using a batch size of $128$ for $100$ epochs. The ADAM optimizer with momentum parameters $(0.9, 0.999)$, and a fixed learning rate of $1e-3$, is used for training. As mentioned in the main paper, for the case of CIFAR-10, we train a ResNet-34 model with the standard training configurations. For the other regression benchmarks, we used a standard MLP with $5$ hidden layers, ReLU activation and batchnorm. They were all trained for $5000$ epochs with learning rate $5e-5$ and ADAM optimizer.

\begin{figure}[h]
    \centering
    \begin{subfigure}[c]{0.99\textwidth}
         \centering
         \includegraphics[width=0.99\textwidth]{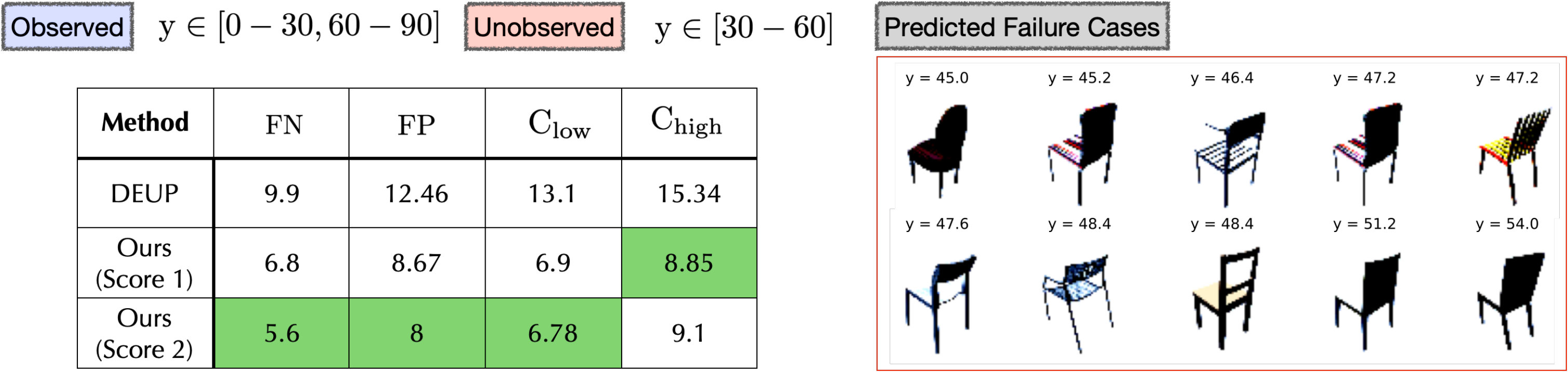}
         \caption{Chair Angles (Gap)}
         \vspace{0.3cm}
         \label{fig:a}
     \end{subfigure}
\begin{subfigure}[c]{0.99\textwidth}         \centering
         \includegraphics[width=0.99\textwidth]{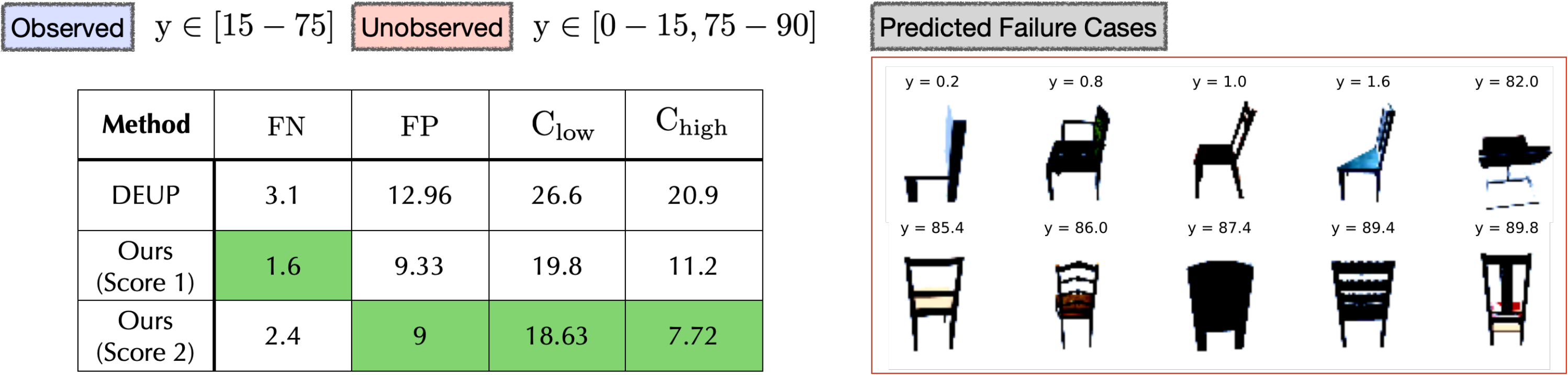}
         \caption{Chair Angles (Tails)}
         \vspace{0.3cm}
         \label{fig:b}
     \end{subfigure}
\begin{subfigure}[c]{0.99\textwidth}         \centering
         \includegraphics[width=0.99 \textwidth]{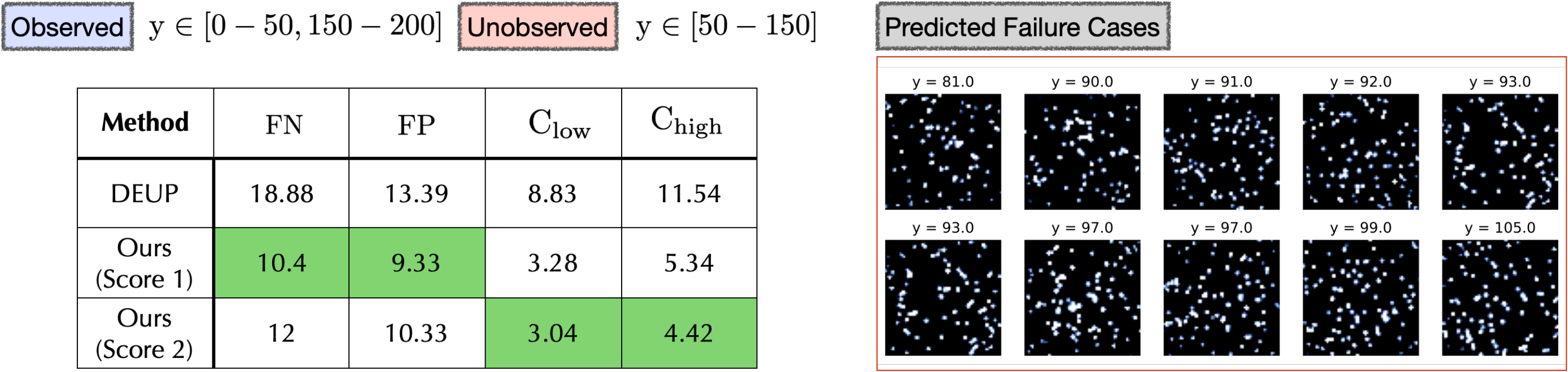}
         \caption{Cells Count (Gap)}
         \vspace{0.3cm}
         \label{fig:c}
     \end{subfigure}
\begin{subfigure}[c]{0.99\textwidth}         \centering
         \includegraphics[width=0.99 \textwidth]{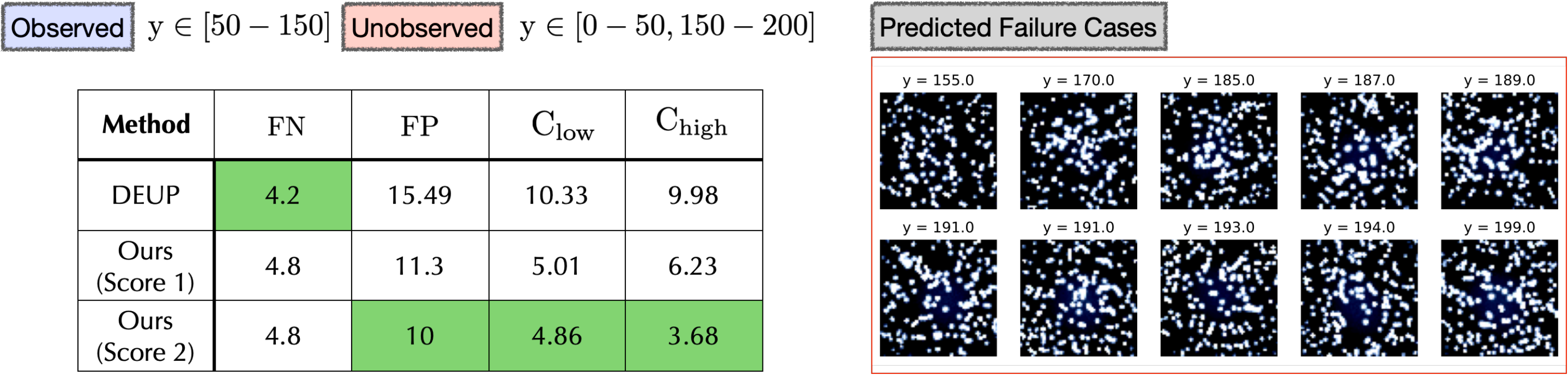}
         \caption{Cells Count (Tails)}
         \vspace{0.3cm}
         \label{fig:d}
     \end{subfigure}
    \caption{\textbf{Efficacy of \name~on Image Regression Benchmarks.} We can observe that in comparison to the state-of-the art baseline DEUP, \name~effectively minimizes the FN, FP and confusion metrics even under challenging extrapolation scenarios. We find that \name~can consistently flag samples from the unobserved regimes which corresponds to highly erroneous predictions.}
\label{fig:image}
\end{figure}

\subsection{Additional Results}
\label{app: 5}
\noindent \textbf{Image regression experiment.} For the cell count and chair angle prediction benchmarks from the main paper, we provide examples of high-risk sample as detected by \name. Please refer to Figure \ref{fig:image} for the examples. Notably, these samples correspond to regimes that were not encountered during training.

\end{document}